\documentclass[3p]{elsarticle}

\usepackage[T1]{fontenc}
\usepackage[caption=false]{subfig}
\usepackage[retain-explicit-plus]{siunitx}
\usepackage{xcolor}
\usepackage{booktabs}
\usepackage{tabu}
\usepackage{tabularx}
\usepackage{tabulary}
\usepackage{multirow,bigdelim}
\usepackage{rotating}
\usepackage{makecell}
\usepackage{threeparttable}
\usepackage{tablefootnote}
\usepackage[ruled,vlined]{algorithm2e}
\usepackage{listings}
\usepackage{enumitem}
\usepackage[hyphens]{url}
\usepackage{lineno,hyperref}
\usepackage[font=normalsize]{caption}
\usepackage{comment}
\usepackage{tablefootnote}

\modulolinenumbers[5]

\journal{Rapid Prototyping Journal}

\biboptions{authoryear}

\usepackage{blindtext}
\usepackage{hyperref}
\usepackage{nameref}

\newcounter{mylabelcounter}

\makeatletter
\newcommand{\labelText}[2]{%
\refstepcounter{mylabelcounter}%
\immediate\write\@auxout{%
  \string\newlabel{#2}{{\unexpanded{#1}}{\thepage}{{\unexpanded{#1}}}{mylabelcounter.\number\value{mylabelcounter}}{}}%
}%
}
\makeatother

\makeatletter
\newcommand\footnoteref[1]{\protected@xdef\@thefnmark{\ref{#1}}\@footnotemark}
\makeatother

\begin{document}

This version of the paper has been accepted for publication after peer review and is available on Emerald Insight at: \url{https://doi.org/10.1108/RPJ-01-2023-0028}\\

\textcopyright This author accepted manuscript is deposited under a Creative Commons Attribution Noncommercial 4.0 International (CC BY-NC) license. This means that anyone may distribute, adapt, and build upon the work for non-commercial purposes, subject to full attribution. If you wish to use this manuscript for commercial purposes, please contact: permissions@emerald.com.
\newpage


\begin{frontmatter}

\title{Toward data-driven research: preliminary study to predict surface roughness in material extrusion using previously published data with Machine Learning}

\author[address1]{Fátima García-Martínez} 
\ead{fatimag4m@gmail.com}

\author[address2]{Diego Carou} \corref{mycorrespondingauthor}
\ead{diecapor@uvigo.es}

\author[address3]{Francisco de Arriba-Pérez}
\ead{farriba@gti.uvigo.es}

\author[address3]{Silvia García-Méndez}
\ead{sgarcia@gti.uvigo.es}

\address[address1]{School of Aeronautics and Space Engineering, University of Vigo, Campus As Lagoas, E32004, Ourense, Spain}

\address[address2]{Design in Engineering Department, University of Vigo, Campus As Lagoas, E32004, Ourense, Spain}

\address[address3]{Information Technologies Group, atlanTTic, University of Vigo, Campus Lagoas-Marcosende, Vigo, 36310, Galicia, Spain}

\cortext[mycorrespondingauthor]{Corresponding author: diecapor@uvigo.es}

\begin{abstract}

\textbf{Purpose.} Material extrusion is one of the most commonly used approaches within the additive manufacturing processes available. Despite its popularity and related technical advancements, process reliability and quality assurance remain only partially solved. In particular, the surface roughness caused by this process is a key concern. To solve this constraint, experimental plans have been exploited to optimize surface roughness in recent years. However, the latter empirical trial and error process is extremely time- and resource-consuming. Thus, this study aims to avoid using large experimental programs to optimize surface roughness in material extrusion.

\textbf{Methodology.} This research provides an in-depth analysis of the effect of several printing parameters: layer height, printing temperature, printing speed and wall thickness. The proposed data-driven predictive modeling approach takes advantage of Machine Learning models to automatically predict surface roughness based on the data gathered from the literature and the experimental data generated for testing.

\textbf{Findings.} Using 10-fold cross-validation of data gathered from the literature, the proposed Machine Learning solution attains a 0.93 correlation with a mean absolute percentage error of 13 \%. When testing with our own data, the correlation diminishes to 0.79 and the mean absolute percentage error reduces to 8 \%. Thus, the solution for predicting surface roughness in extrusion-based printing offers competitive results regarding the variability of the analyzed factors.

\textbf{Originality.} Although Machine Learning is not a novel methodology in additive manufacturing, the use of published data from multiple sources has barely been exploited to train predictive models. As available manufacturing data continue to increase on a daily basis, the ability to learn from these large volumes of data is critical in future manufacturing and science. Specifically, the power of Machine Learning helps model surface roughness with limited experimental tests.

\textbf{Research limitations.} There are limitations in obtaining large volumes of reliable data, and the variability of the material extrusion process is relatively high. 

\end{abstract}


\end{frontmatter}

\section{Introduction}

Nowadays, data are a pivotal asset for the world economy. In fact, the term data economy is becoming popular \citep{Carvalho2022}. Large amounts of data are created, captured, copied and consumed every year. Consequently, the IDC Global Data Sphere report established the size of global data in 59 zettabytes as a baseline \citep{zillner2021data}. Science is aware of the shift toward more data-oriented research. The increasing availability of data creates new possibilities to develop more accurate and reliable models that may help verify or reject research hypotheses. However, it is still not easy to access most data or to even obtain properly curated data to be used \citep{hey2009fourth}. Mainly, scientific research combines both open and sub-scripted books, conference proceedings and journals. However, in science, there is no common standard for arranging data for publication, and there are times when data are simply missing from the studies.

Traditionally, linear regression models have largely been used in research to investigate the causal relationship between multiple independent variables and dependent variables \citep{efendi2023cleansing}. However, these models are based on strong assumptions, such as normally distributed errors, which may be violated when large data sets from different sources are used \citep{Thongpeth2021}. Therefore, these traditional statistical approaches are not good enough when it comes to addressing big data issues. The Machine Learning (\textsc{ml}) models are good alternatives to linear regression without the normality assumption (\textit{i.e.}, when the relationship between the outcome and its determinants is non-linear) \citep{Thongpeth2021}. In this sense, artificial intelligence allows patterns to be identified in data by training computers \citep{shu2022knowledge}. Over the last few decades, there have been great advances in the artificial intelligence field. Applications of various levels of complexity have been developed for a wide range of industrial sectors to solve very different types of problems. For example, the automotive, fishing, food and beverage industries are introducing these techniques \citep{caroumachine2022}. In fact, advanced predictive modeling is essential to ensure high-quality products and to further optimize processes. Particularly, advancements in data analytics \citep{Awan2021} will benefit the smart manufacturing sector \citep{Cui2020,caroumachine2022}.

In engineering, research efforts are mainly conducted by research labs that develop experimental research using their own means. Despite the fact that these efforts are connected to the literature, some studies are developed around the same hypothesis and methodologies, making the conclusions redundant. Additionally, some efforts may not be reproducible and, therefore, are of little value. According to \cite{kozlov2023disruptive}, science in the twenty-first century is less disruptive than science in the mid-twentieth century. Moreover, research is more oriented toward obtaining incremental advances. This explanation is based on a study of a large set of manuscripts (45 million) and patents (3.9 million), which states that less departure from previous literature results in less disruptive research. However, although the research is based on the literature, the large quantity of published research makes it impossible for researchers to have comprehensive and accurate knowledge of the current state-of-the-art due to the lack of capabilities and resources to process this information. In other words, researchers may not know some key data, verified hypotheses, and general know-how. In contrast, when provided with huge amounts of data, artificial intelligence can obtain knowledge from those sources to help develop new adequate research projects.

Presently, additive manufacturing is being extensively used in the automotive and electronic industries \citep{Leal2017,Saengchairat2017,Bockin2019} and in healthcare products \citep{Zadpoor2017,RezvaniGhomi2021}, among many other cases. Its economic and flexible nature stands out as a relevant advantage \citep{Ford2016}. It can be classified into seven processes \citep{Wu2019}: (\textit{i}) binder jetting, (\textit{ii}) directed energy deposition, (\textit{iii}) material extrusion, (\textit{iv}) material jetting, (\textit{v}) powder bed fusion, (\textit{vi}) vat polymerization, and (\textit{vii}) sheet lamination. 

Material extrusion was first intended for prototyping applications though its potential enables the creation of fully functional products. This process is of special interest because it is the most commonly used approach. This is due to the existence of affordable solutions and the variety of usable materials (mainly plastic) \citep{Messimer2019}. 

Unfortunately, even though the use of material extrusion is increasing and becoming more mature, process reliability and quality assurance remain an open challenge \citep{Baumers2017}. During the last few decades, many experimental studies have been presented to optimize the printing parameters that might have an effect on the printing results \citep{Panda2015,Panda2017}. Specifically, material extrusion processes have been thoroughly revised to enhance surface quality \citep{Perez2018}, form (cylindrical, staircase error, flatness and straightness of parts) \citep{Haghighi2018} and mechanical resistance, among others \citep{Moreno2021}. For instance, printed parts are subjected to the staircase effect, which depends on the layer height and face inclination \citep{Paul2015}. Issues such as the staircase effect can be limited by diminishing the layer height \citep{Gupta2022}. However, the generalizability of these parameter optimization studies is limited regarding dimensional accuracy. According to a study by \cite{golab2022generalisable} "The variety of optimal printing parameter values suggests that unique solutions exist for each combination of machine, material, and artefact". 

Among the studies on quality assurance, both the research community and related enterprises have considered surface finish \citep{Wu2019}. In particular, surface roughness is a critical constraint that usually comes as a specification in engineering products \citep{Demircioglu2013}. In light of this, analyzing the influence of the processing conditions becomes crucial to meet the part's requirements. In the specific case of surface roughness, the lower, the better. 

Different approaches have been used to analyze surface roughness in material extrusion, including analytical \citep{chohan2016mathematical,Lalehpour2018,EhsanulHaque2019,Wang2019} and experimental \citep{Perez2018}. However, very limited effort has been directed to \textsc{ml} approaches, particularly for studies using published data. Based on all of the above, it is hypothesized that by using the available published data, a model can be developed to predict outputs in manufacturing processes without needing to perform large, costly and time-consuming experimental plans. 

In this study, the objective is to focus on the material extrusion process. In this sense, the study is developed as a preliminary analysis to predict surface quality in material extrusion by utilizing \textsc{ml}. Specifically, this article aims to analyze the influence of layer height, printing speed, printing temperature and wall thickness on surface roughness using published data and experimental data. Consequently, these models might demonstrate that surface roughness can be modeled avoiding the need to perform large experimental programs by taking advantage of the aforementioned data from the literature.

The article is organized in the following way. Section \ref{sec:related} reviews related work on predictive modeling in material extrusion, considering both traditional \textsc{ml} and Deep Learning (\textsc{dl}) models for surface roughness prediction. Section \ref{sec:method} describes our novel proposal. Section \ref{sec:results} presents the experimental results. Finally, Section \ref{sec:conclusion} identifies the main conclusions and Section \ref{sec:limitations and future work} presents the main limitations of the study and insights for future research.

\section{Related work}
\label{sec:related}

A variety of factors exist that directly affect the printing process and can be controlled programmatically (\textit{e.g.}, layer height). Accordingly, the Ultimaker Cura printer, one of the most popular software for 3\textsc{d} printers, offers over 400 different settings in the custom mode. However, some factors cannot be controlled before the printing process takes place (\textit{e.g.}, vibrations).

In the specific case of surface roughness, considerable effort has been carried out to identify the potential factors that most affect printed parts. These efforts have resulted in a generally accepted body of knowledge about the most important factors that might affect surface roughness: bed temperature, fan speed, infill density and pattern, layer height, printing speed and temperature, type and state of the material, type of printer, and wall line count \citep{Bikas2016,Hallgren2016,bourell2017materials,Umaras2017,Perez2018,Li2019}. This background was mainly gathered by testing the experimental designs developed by the research community. In general, the experimental studies confirm that layer height is critical for surface roughness \citep{Perez2018}, but the effect of other printing parameters on surface roughness is not well-known yet. For instance, \cite{Mendricky2020} identified the influence of printing speed on surface roughness, whereas \cite{Yang2020} found that this factor did not influence this output. Consequently, there is still a research gap that requires more testing to validate published findings.

Among data mining research applied to additive manufacturing, traditional approaches \citep{Barrios2019,Li2019,Li2019theoretical,Wu2019,Liu2021} and \textsc{dl} or Neural Networks (\textsc{nn}) approaches \citep{Hooda2021,Kandananond2021,Tripathi2021,Saad2022} deserve attention. However, the use of the former \textsc{nn} models is limited due to the huge amount of training data needed. 

\cite{Barrios2019} proposed a \textsc{ml}-based solution for surface roughness in material extrusion. Specifically, the authors exploited the Decision Tree (\textsc{dt}) and Random Forest (\textsc{rf}) models, along with J48 and Random Tree (\textsc{rt}) combined with the L27 array-based Taguchi method. The outcome allows the end users to efficiently and effectively set the flow rate, layer height, print acceleration, speed and temperature parameters.

\cite{Li2019} presented a more sophisticated data-driven solution with an ensemble \textsc{ml} model to predict surface roughness in two stages (offline and online). For the offline operation mode, classification and regression trees were used to train the system. Afterward, the testing stage was performed in online mode. Note that the surface roughness was estimated in a unique direction. Moreover, the MakerBot Replicator Plus 3\textsc{d} printer was used to obtain the parts. \cite{Wu2019} presented a similar \textsc{ml}-based work, and the prediction accuracy was subsequently enhanced by \cite{Li2019theoretical}. Particularly, they based their model on the parabolic curve and linear straight-line surface profile representation method.

More recently, \cite{Hooda2021} used \textsc{ml} models to predict the optimum deposition angle. The training data were obtained from the Ultimaker Cura software using different geometries for the Prusa i3 printer. The authors used Linear and Logistic Regression (\textsc{lr}), \textsc{dt}, \textsc{rf} and Support Vector Machine (\textsc{svm}) models, along with Multi-layer Perceptron and \textsc{nn}. In the end, the \textsc{rf} classifier offered the best results. \cite{Kandananond2021} also used Multi-layer Perceptrons. More in detail, they applied two training methods (weight backtracking-resilient backpropagation and globally convergent resilient backpropagation) to compare the effect of the number of neurons in hidden layers.

\cite{Tripathi2021} analyzed the impact of several parameters on surface roughness using \textsc{dl}. However, the authors did not use K-fold cross-validation, leaving this method for future work, and admitted the need for larger amounts of data. Similarly, \cite{Jiang2022} presented a theoretical solution for surface roughness in material-extruded products. The authors concluded that extrusion temperature and layer thickness are the most relevant factors. Finally, although the use of \textsc{nn} is limited due to the huge amount of training data needed, \cite{Saad2022} analyzed the influence of several network structures on the printing process, using the coefficient of determination as a representative parameter in their study.

More closely related to our work, \cite{Liu2021} presented a detailed study on additive manufacturing to endorse the application of \textsc{ml} models. Their study is based on data gathered from 121 articles in the literature on powder laser-based additive manufacturing of the Ti-6Al-4V alloy. When dealing with data mining, the authors identified some important drawbacks, such as: “[...]these data do not consistently record composition, micro-structure, thermal history, part shape and orientation, starting powder composition and density, and other important process and material properties[...]”. Despite the limitations, the authors proved that \textsc{ml} can be used as a starting point for further analysis.

The ability to learn from these large volumes of data will be critical in future manufacturing and science. The traditional approaches require a lot of testing and are highly demanding. To the best of our knowledge, this is the first research study to combine multi-source data sets and \textsc{ml} techniques to train a surface roughness predictive model in material extrusion. The main objective is to combine data from the literature in order to (i) reduce time and resource constraints, and (ii) analyze the coherence of the literature data from different printing machines.

\section{Materials and methods}
\label{sec:method}

\begin{figure*}[!htbp]
\centering
\includegraphics[scale=0.17]{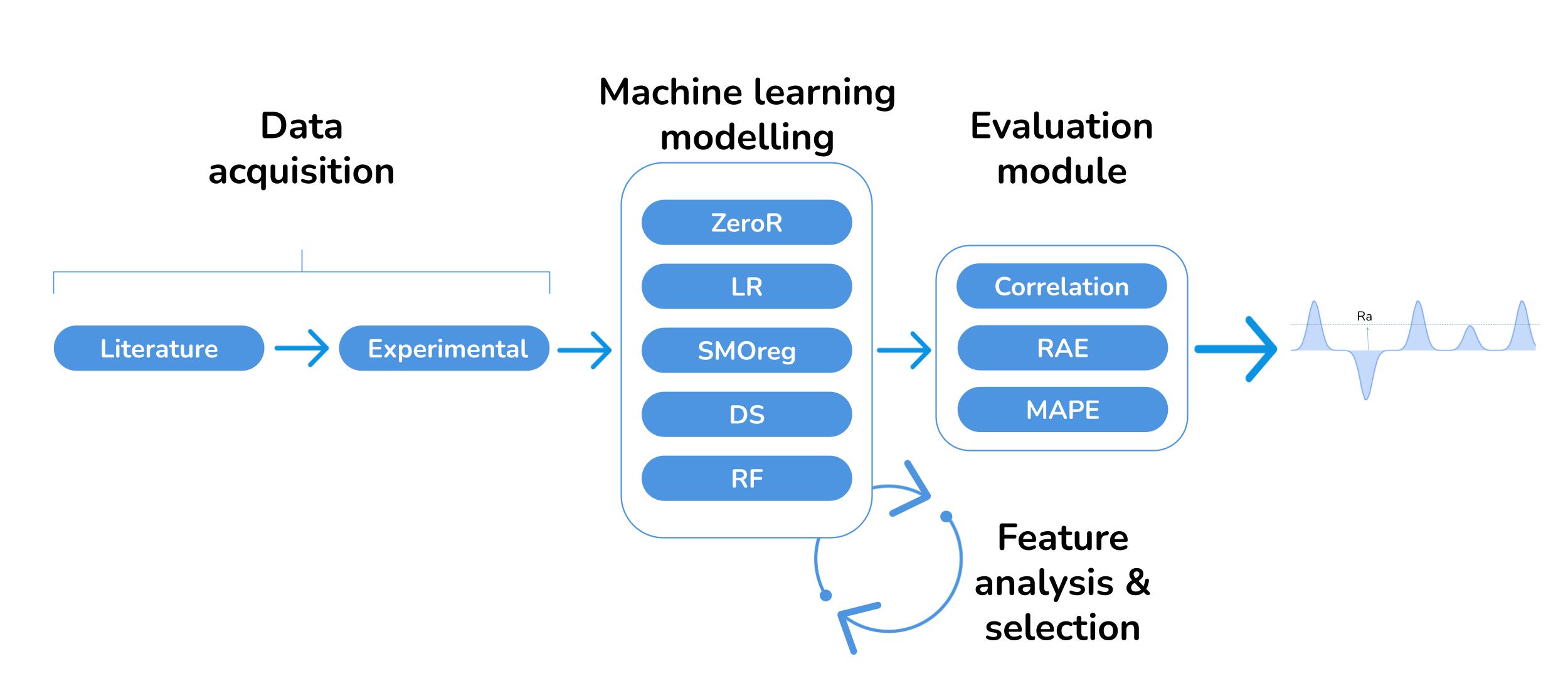}
\caption{\label{fig:solution} System scheme (figure by authors).}
\end{figure*}

Figure \ref{fig:solution} details the system scheme developed for the surface roughness prediction experimental plan, which it is composed of three stages: (\textit{i}) data acquisition (Section \ref{sec:data_acquisition} and \ref{sec:experimental_plan}); (\textit{ii}) \textsc{ml} modeling (Section \ref{sec:ml_modelling}), including feature analysis and selection processes (Section \ref{sec:feature_analysis_selection}); and (\textit{iii}) evaluation (Section \ref{sec:evaluation_metrics}). In Section \ref{sec:implementations}, the computer requirements and implementations are described.

Initially, data acquisition was performed to create an adequate and reliable experimental data set. Afterward, an optimal \textsc{ml} model was created to predict the surface roughness based on the training data set. In the end, data from experimental tests were used to validate the model.

\subsection{Data acquisition}
\label{sec:data_acquisition}

Data were obtained from published experimental studies in the literature \citep{Alsoufi2017,Peng2018,aslani2020quality,Buj-Corral2021,Morampudi2021,Tasdemir2021,Vambol2021}. Several manuscripts were discarded based on criteria, such as not indicating the measuring surface or only presenting graphs rather than the recorded values. Furthermore, a testing data set was obtained by experimental testing. The former data are composed of 100 training values, whereas the latter data are composed of 128 experimental values. Note that the final data set\footnote{\label{data_link}Available at \url{https://bit.ly/3RIEA8i}, April 2023.} is similar in size to other state-of-the-art studies \citep{Gu2018,jiang2022superior,Westphal2022}.

In this study, the experimental data set (used for testing) is 28 \% larger than the merged data set retrieved from the literature (used for training). However, the latter smaller data set includes data related to different printers; therefore, different printing patterns will contribute to creating a general \textsc{ml} model. The ultimate objective is to validate whether the general data set from the literature is able to predict the surface roughness of a different printer in different laboratory environments.

\subsection{Experimental plan}
\label{sec:experimental_plan}

To create a data set to validate the models, an experimental plan consisting of 16 tests was defined. The samples were printed with four varying factors. The combinations were selected with the objective to analyze these four factors with a limited number of samples. The printing tests were performed using polylactic acid (\textsc{pla})), which is characterized by the data listed in Table \ref{tab:characterization_pla}. The material selection helps to model the outcome by taking into account that different materials require different printing parameters, such as bed temperature and printing temperature. 

The printed samples consisted of cylindrical pieces of 30 mm diameter and 60 mm height that were printed via the Ultimaker S5 (referred to as printer \textsc{us}5) and the Ultimaker 3 extended (referred to as printer \textsc{u}3) using the Ultimaker Cura 4.13.1 software. The resolution of the two printers was 20-200 µm. The nozzle had a diameter of 0.4 mm. The factors analyzed included the following: layer height, printing speed, printing temperature and wall thickness. In addition, the infill density was 20 \%. Thus, two sets of eight samples were printed with each of the printers. The full breakdown can be seen in Table \ref{tab:printing_conditions}.

\begin{table*}[!htbp]
\centering
\caption{\label{tab:characterization_pla} Characteristics and recommended processing conditions of the \textsc{pla} (table by authors).}
\begin{tabular}{ll}
\toprule
 \textbf{Composition} & \textsc{pla} (polylactic resin) - 99 \% CAS: 9051-89-2\\
 \textbf{Density} & 1.24 g/cm$^3$\\
 \textbf{Diameter} & 1.75 ± 0.03 mm\\
 \textbf{Printing temperature} & 220 ± 20 ºC\\
 \textbf{Melting temperature} & 180 ºC
\\\bottomrule
\end{tabular}
\end{table*}

\begin{figure*}[!htbp]
\centering
\includegraphics[scale=0.6]{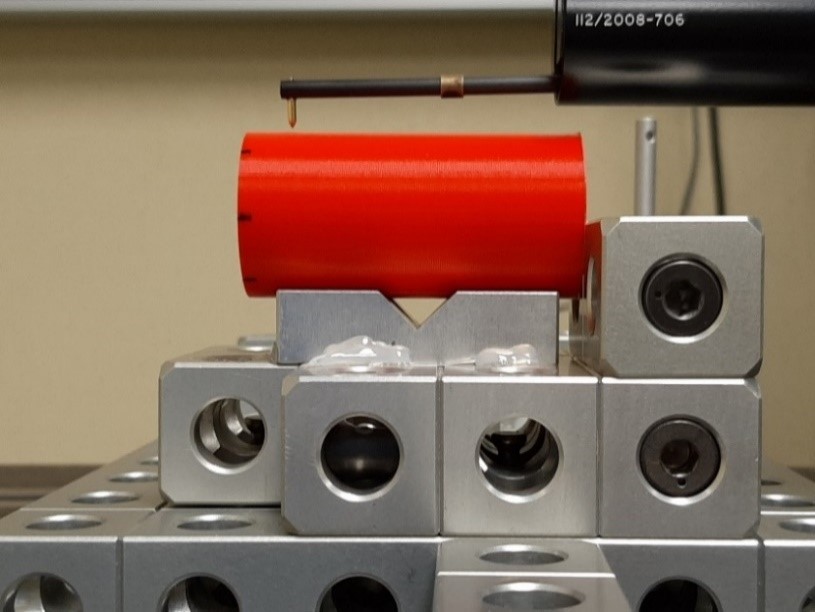}
\caption{\label{fig:profilometer} Surface roughness profilometer (figure by authors).}
\end{figure*}

\begin{table}[ph!]
\centering
\caption{\label{tab:printing_conditions}{Factors and levels in the experimental plan (table by authors).}}
\begin{tabular}{ccccccc}
\toprule
\bf Sample & \bf Printer & \bf Layer & \bf Printing & \bf Printing & \bf Wall & \bf Infill\\

& & \textbf{height} (mm) & \textbf{speed} (mm/s) & \textbf{temperature} (°C) & \textbf{thickness} (mm) & \textbf{density} (\%) \\

\midrule
1 & \textsc{us}5 & 0.15 & 60 & 200 & 2 & 20\\
2 & \textsc{us}5 & 0.23 & 60 & 200 & 2 & 20\\
3 & \textsc{us}5 & 0.30 & 60 & 200 & 2 & 20\\
4 & \textsc{us}5 & 0.23 & 60 & 215 & 2 & 20\\
5 & \textsc{us}5 & 0.23 & 60 & 230 & 2 & 20\\
6 & \textsc{us}5 & 0.23 & 60 & 200 & 1 & 20\\
7 & \textsc{us}5 & 0.23 & 60 & 200 & 3 & 20\\
8 & \textsc{us}5 & 0.23 & 40 & 200 & 2 & 20\\
9 & \textsc{u}3 & 0.15 & 60 & 200 & 2 & 20\\
10 & \textsc{u}3 & 0.23 & 60 & 200 & 2 & 20\\
11 & \textsc{u}3 & 0.30 & 60 & 200 & 2 & 20\\
12 & \textsc{u}3 & 0.23 & 60 & 215 & 2 & 20\\
13 & \textsc{u}3 & 0.23 & 60 & 230 & 2 & 20\\
14 & \textsc{u}3 & 0.23 & 60 & 200 & 1 & 20\\
15 & \textsc{u}3 & 0.23 & 60 & 200 & 3 & 20\\
16 & \textsc{u}3 & 0.23 & 40 & 200 & 2 & 20\\
\bottomrule
\end{tabular}
\end{table}

\begin{figure*}[!htbp]
\centering
\includegraphics[scale=0.08]{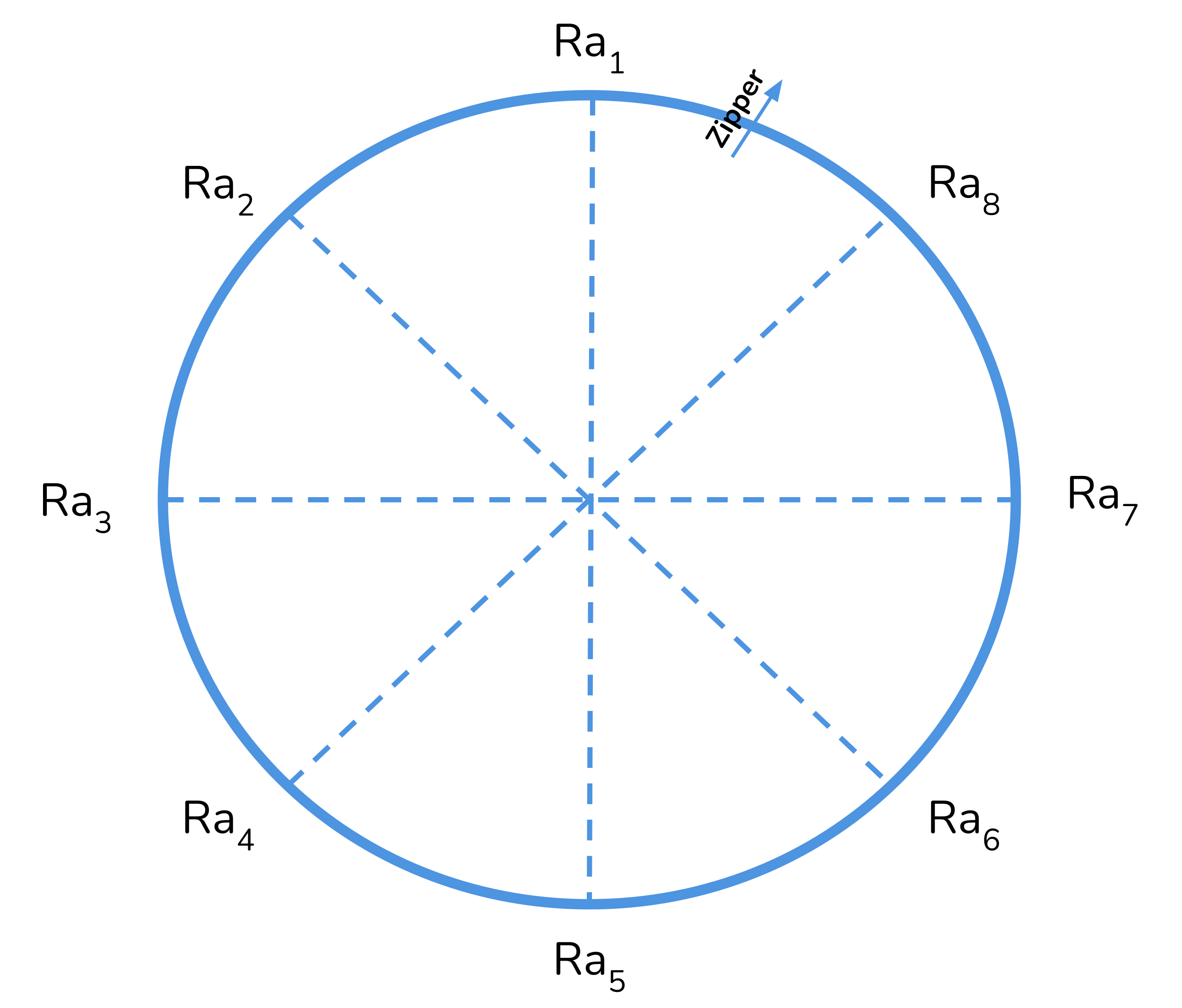}
\caption{\label{fig:measures_ra} Measuring position and location of the zipper (top view of the printing bed) (figure by authors).}
\end{figure*}

The arithmetic average roughness (\textit{Ra}) was computed as the response variable. The ISO 4287 standard was used to record the roughness using five cut-offs of 8 mm. Thus, a length of 40 mm was evaluated because the expected surface roughness values were higher than 10 $\mu$m. To measure the surface roughness, a Taylor Hobson profilometer was used (see Figure \ref{fig:profilometer}). The surface roughness measurement was performed in eight different locations for each sample. Figure \ref{fig:measures_ra} details the eight positions with a 45º separation. The average surface roughness at these eight positions is defined as \textit{Ra1}, \textit{Ra2}, \textit{Ra3}, \textit{Ra4}, \textit{Ra5} \textit{Ra6}, \textit{Ra7}, and \textit{Ra8}. 

\subsection{Machine Learning modeling}
\label{sec:ml_modelling}

Since the literature data are collected from different printers and acquisition environments, the correlation between the layer height feature and the target feature in the literature merged data set (0.82 in Table \ref{tab:corr_literature}) can be used as a baseline and motivation to apply a more sophisticated multi-modal approach to extract patterns with greater precision. {Table \ref{tab:corr_literature} shows the linear correlations between all of the features in the study and the target feature, regardless of the model used.} Accordingly, different \textsc{ml} models were selected to evaluate the data set using diverse analysis perspectives.

\begin{table*}[!htbp]
\centering
\caption{\label{tab:corr_literature}Correlation of features in the data set extracted from the literature (table by authors).}
\begin{tabular}{lc}
\toprule
 \textbf{Printing parameter} & \textbf{Significance}\\\hline
 Layer height & 0.82\\
 Nozzle diameter & 0.35\\
 Wall thickness & 0.16\\
 Infill density & 0.09\\
 Printing temperature & 0.06\\
 Printing speed & -0.03\\
 Shape & -0.21
\\\bottomrule
\end{tabular}
\end{table*}

The models were trained and tested with 10-fold cross-validation \citep{Berrar2019} using the data from the literature exclusively. The latter process consists of using different parts of the experimental data set over multiple iterations for both training and testing models. Consequently, it is possible to estimate the precision with which a model will be able to predict new samples in a real scenario. The latter process also contributes to minimizing data selection randomization during evaluation. In the end, the proposed model is validated using our own data gathered throughout this project.

Two operation modes in \textsc{ml}-based solutions exist that depend on the feasible values of the target feature: (\textit{i}) nominal and (\textit{ii}) categorical. This work belongs to the first group. Thus, the \textsc{ml} models seek to infer the best mathematical equation to fit the training data acting as regressor models. Motivated by their good performance in related studies in the literature \citep{Barrios2019,Hooda2021}, the following models were selected:

\begin{itemize}
 \item \textbf{ZeroR} \citep{Sangeorzan2020} is a rule-based model that operates with the mean of the data.
 
 \item \textbf{\textsc{lr}} \citep{Pal2019} uses the Akaike criterion for model selection and is capable of handling weighted instances.
 
 \item \textbf{\textsc{smo}reg} \citep{Assim2020} implements a specific \textsc{svm} for regression problems.
 
 \item \textbf{Decision Stump} (\textsc{ds}) \citep{Uskov2021} belongs to the group of tree algorithms and is one of the most basic.
 
 \item \textbf{\textsc{rf}} \citep{Schonlau2020} belongs to the group of tree algorithms, but follows a more complex approach than \textsc{ds}. Particularly, it is composed of a set of $n$ trees that finally ends up assembled into one to make a prediction.
\end{itemize}

For the \textsc{ml} models, we selected implementations from Weka \textsc{ml} software in Java: ZeroR\footnote{Available at \url{https://weka.sourceforge.io/doc.stable-3-8/weka/classifiers/rules/ZeroR.html}, April 2023.}, \textsc{lr}\footnote{Available at \url{https://weka.sourceforge.io/doc.dev/weka/classifiers/functions/LinearRegression.html}, April 2023.}, \textsc{smo}reg\footnote{Available at \url{https://weka.sourceforge.io/doc.dev/weka/classifiers/functions/SMOreg.html}, April 2023.}, \textsc{ds}\footnote{Available at \url{https://weka.sourceforge.io/doc.dev/weka/classifiers/trees/DecisionStump.html}, April 2023.}, and \textsc{rf}\footnote{Available at \url{https://weka.sourceforge.io/doc.dev/weka/classifiers/trees/RandomForest.html}, April 2023.}.

The aforementioned models are executed with default parameters. Based on the results in Table \ref{tab:corr_literature}, the correlation between the input features and the surface roughness target will be higher in the \textsc{ml} models and will have more complex relationships among them, ensuring competing results. Furthermore, for both the experimental data and the data gathered from the literature, the data variability is assumed to be minimal. Otherwise, the \textsc{ml} models will not converge, and the results will be extremely low.

\subsubsection{Feature analysis and selection}
\label{sec:feature_analysis_selection}

This module exhibits a circular behavior (\textit{i.e.}, first, the features are optimized, and then, the model is re-evaluated). Using feature ranking techniques, a list of the most relevant features is obtained, excluding those that negatively influence the model. Afterward, the \textsc{ml} models are evaluated. Note that the feature selection seeks to (\textit{i}) eliminate redundant features, (\textit{ii}) improve the results, and (\textit{iii}) reduce the complexity of the model in terms of the number of features used.

\subsection{Evaluation module}
\label{sec:evaluation_metrics}

When \textsc{ml} models act as classifiers, they are traditionally evaluated using metrics such as accuracy, precision and recall, because of the limited number of values available to predict the target feature. However, when they are used as regressors, as in this study, the unlimited target values support the use of correlation analysis and relative absolute error (\textsc{rae}, see Equation \ref{eq:rae}) where $f_i$ and $a_i$ correspond to the forecast and actual values, respectively. 

\begin{equation}
\label{eq:rae}
    RAE  (\%) = 100\frac{\sum_{i=1}^{n} |f_i-a_i|}{\sum_{i=1}^{n} |\overline{a}-a_i|}
\end{equation}

The former metric assesses the model's adaptability to the target feature variations (decreases and increases). The \textsc{rae} metric evaluates the predicted value's average-percentual deviation from the expected one. However, in a problem with sparse target values, such as this case (the surface roughness values range from 3.88 to 32.79 $\mu$m for the training data, and in the experimental data, from 8.53 to 27.02 $\mu$m), the \textsc{rae} metric may provide rather despairing results. The latter is due to the fact that the \textsc{rae} metric weighs using the average of the actual values of surface roughness. Consequently, the mean absolute percentage error (\textsc{mape}) metric was also computed to perform a much more precise evaluation of the \textsc{ml} models (see Equation \ref{eq:mape}). \textsc{mape} metric is a measure of the error between forecast and actual values with respect actual values, whereas the \textsc{rae} metric measures the error with respect to the deviation of mean actual values. Thus, based on similar works in the literature \citep{Rahmati2015,Vahabli2016}, \textsc{mape} is the most appropriate metric to assess the surface roughness prediction model.

\begin{equation}
\label{eq:mape}
    MAPE (\%) = \frac{100}{n} \sum_{i=1}^{n} |\frac{f_i-a_i}{a_i}| 
\end{equation}

The tolerance error (\textsc{te}) shown by the printers must also be taken into account as the lower limit. Accordingly, \textsc{te} values have been reported to oscillate between 10 and 40 \% in the literature \citep{Kairn2021}. In the end, combining the aforementioned metrics allows us to assess the effectiveness of the proposed solution from a complete perspective.

\subsection{Experimental setting and implementations}
\label{sec:implementations}

The experiments were performed on a computer with the following specifications:

\begin{itemize}
 \item Operating system: Ubuntu 18.04.2 \textsc{lts} 64 bits
 \item Processor: Intel\@Core i9-9900K 3.60 GHz 
 \item RAM: 32 GB DDR4 
 \item Disk: 500 GB (7200 rpm SATA) + 256 GB SSD
\end{itemize}

\section{Results and discussion}
\label{sec:results}

The experimental results and the development of the \textsc{ml} models are described; and the results obtained and the limitations of the proposed method are discussed.

\subsection{Machine Learning modeling}

Table \ref{tab:roughness_results} presents the surface roughness results obtained for the 16 samples manufactured with the aforementioned printing conditions. Note the variability in the surface roughness values. For the same sample and printer, the surface roughness can reach values up to 27.02 $\mu$m (see sample 14). Moreover, compared to the Ultimaker S5, the Ultimaker 3 extended printer attains up to 8.52 $\mu$m more in average surface roughness (see the comparison between samples 6 and 14 in Table \ref{tab:roughness_comparison}).

\begin{table*}[!htbp]
\centering
\caption{\label{tab:roughness_results} {Surface roughness results (table by authors).}}
\begin{tabular}{cccccccccc}
\toprule
\bf Sample & \begin{tabular}[c]{@{}c@{}}\bf \textit Ra1 \\ (mm)\end{tabular} & \begin{tabular}[c]{@{}c@{}}\bf \textit Ra2 \\ (mm)\end{tabular} & \begin{tabular}[c]{@{}c@{}}\bf \textit Ra3 \\ (mm)\end{tabular} & \begin{tabular}[c]{@{}c@{}}\bf \textit Ra4 \\ (mm)\end{tabular} & \begin{tabular}[c]{@{}c@{}}\bf \textit Ra5 \\ (mm)\end{tabular} & \begin{tabular}[c]{@{}c@{}}\bf \textit Ra6 \\ (mm)\end{tabular} & \begin{tabular}[c]{@{}c@{}}\bf \textit Ra7 \\ (mm)\end{tabular} & \begin{tabular}[c]{@{}l@{}}\bf \textit Ra8 \\ (mm)\end{tabular} & \begin{tabular}[c]{@{}c@{}}\bf Avg \textit{Ra} \\ (mm)\end{tabular} \\\midrule
1 & 8.64 & 9.39 & 9.22 & 9.84 & 9.70 & 8.53 & 8.84 & 9.28 & 9.18 \\
2 & 15.70 & 15.95 & 15.83 & 16.63 & 16.34 & 16.26 & 16.22 & 16.33 & 16.16 \\
3 & 22.61 & 22.76 & 22.33 & 21.89 & 20.87 & 22.17 & 22.11 & \bf 22.52 & 22.03 \\
4 & 15.28 & 15.64 & 15.86 & 16.77 & 14.98 & 15.16 & 15.95 & 16.15 & 15.72 \\
5 & 16.29 & 16.35 & 16.53 & 15.51 & 16.99 & 15.26 & 15.91 & 16.55 & 16.17 \\
6 & 15.70 & 15.52 & 15.53 & 14.81 & 16.58 & 15.82 & 15.37 & 15.82 & 15.64 \\
7 & 15.14 & 15.44 & 14.93 & 15.76 & 15.83 & 14.48 & 15.61 & 15.51 & 15.34 \\
8 & 15.68 & 16.30 & 16.19 & 17.01 & 15.95 & 15.30 & 16.62 & 16.18 & 16.15 \\
9 & 10.19 & 11.79 & 11.29 & 11.87 & 10.90 & 10.84 & 9.55 & 10.53 & 10.87 \\
10 & 17.62 & 18.57 & 19.74 & 18.14 & 16.10 & 16.81 & 16.79 & 17.29 & 17.63 \\
11 & 21.65 & 21.90 & 21.98 & 22.00 & 21.14 & \bf 22.08 & \bf 21.99 & 21.41 & 21.77 \\
12 & 17.10 & 16.36 & 17.65 & 18.44 & 16.61 & 18.31 & 15.66 & 18.27 & 17.30 \\
13 & 17.55 & 18.84 & 17.60 & 17.26 & 16.06 & 15.76 & 15.55 & 17.73 & 16.99 \\
14 & \bf 26.49 & \bf 27.02 & \bf 26.25 & \bf 26.00 & \bf 22.70 & 21.57 & 21.47 & 21.81 & \bf 24.16 \\
15 & 17.17 & 17.31 & 18.31 & 16.97 & 16.11 & 17.44 & 16.25 & 17.42 & 17.12 \\
16 & 18.09 & 18.72 & 19.03 & 18.53 & 17.09 & 16.89 & 17.80 & 18.97 & 18.14 \\
\bottomrule
\end{tabular}
\end{table*}

\begin{table*}[!htbp]
\centering
\caption{\label{tab:roughness_comparison} {Comparison of surface roughness results (in absolute value) between the Ultimaker S5 and the Ultimaker 3 extended printers (table by authors).}}
\begin{tabular}{cccccccccc}
\toprule
\begin{tabular}[c]{@{}c@{}}\bf Compared \\ \bf samples\end{tabular} & \begin{tabular}[c]{@{}c@{}}\bf $\Delta$ \textit Ra1 \\ (mm)\end{tabular} & \begin{tabular}[c]{@{}c@{}}\bf $\Delta$ \textit Ra2 \\ (mm)\end{tabular} & \begin{tabular}[c]{@{}l@{}}\bf $\Delta$ \textit Ra3 \\ (mm)\end{tabular} & \begin{tabular}[c]{@{}c@{}}\bf $\Delta$ \textit Ra4 \\ (mm)\end{tabular} & \begin{tabular}[c]{@{}c@{}}\bf $\Delta$ \textit Ra5 \\ (mm)\end{tabular} & \begin{tabular}[c]{@{}c@{}}\bf $\Delta$ \textit Ra6 \\ (mm)\end{tabular} & \begin{tabular}[c]{@{}l@{}}\bf $\Delta$ \textit Ra7 \\ (mm)\end{tabular} & \begin{tabular}[c]{@{}c@{}}\bf $\Delta$ \textit Ra8 \\ (mm)\end{tabular} & \begin{tabular}[c]{@{}c@{}}\bf $\Delta$ Avg \textit{Ra} \\ (mm)\end{tabular} \\\midrule

1, 9 & 1.55 & 2.40 & 2.07 & 2.03 & 1.20 & 2.31 & 0.71 & 1.25 & 1.69 \\
2, 10 &1.92 & 2.62 & 3.91 & 1.51 & 0.24 & 0.55 & 0.57 & 0.96 & 1.47 \\
3, 11 &0.96 & 0.86 & 0.35 & 0.11 & 0.27 & 0.09 & 0.12 & 1.11 & 0.26 \\
4, 12 &1.82 & 0.72 & 1.79 & 1.67 & 1.63 & 3.15 & 0.29 & 2.12 & 1.58 \\
5, 13 &1.26 & 2.49 & 1.07 & 1.75 & 0.93 & 0.50 & 0.36 & 1.18 & 0.82 \\
6, 14 & 10.79 & 11.50 & 10.72 & 11.19 & 6.12 & 5.75 & 6.10 & 5.99 & \bf 8.52 \\
7, 15 &2.03 & 1.87 & 3.38 & 1.21 & 0.28 & 2.96 & 0.64 & 1.91 & 1.78 \\
8, 16 &2.41 & 2.42 & 2.84 & 1.52 & 1.14 & 1.59 & 1.18 & 2.79 & 1.99\\
\bottomrule
\end{tabular}
\end{table*}

Table \ref{tab:ml_literature} details the classification results obtained with 10-fold cross-validation. The correlation value of the best classifier, \textsc{rf}, is 0.93 {(0.11 higher than the baseline established in Table \ref{tab:corr_literature})}. Although the \textsc{rae} reaches 30.16 \%, the \textsc{mape} is 13.01 \%. The latter value means that the surface roughness prediction presents an accuracy of 87 \%. These values are considered competitive taking into account that data come from different sources (\textit{i.e.}, different printing environments, materials and printers). For the rest of the classifiers, especially in the case of the \textsc{smo}reg and \textsc{lr} models, the results obtained are close to these values, which endorses the steady nature of the final data set.

\begin{table*}[!htbp]
\centering
\caption{\label{tab:ml_literature}Correlation and error values for the selected \textsc{ml} models using the data sets from the literature for training (table by authors).}
\begin{tabular}{lllllll}
\toprule
 & \textbf{\textsc{z}ero\textsc{r}} & \textbf{\textsc{lr}} & \textbf{\textsc{smo}reg} & \textbf{\textsc{ds}} & \textbf{\textsc{rf}} \\\hline
 \bf Correlation & -0.28 & 0.90 & 0.91 & 0.74 & 0.93\\
 \bf \textsc{rae} (\%) & 100.00 & 38.50 & 36.24 & 58.27 & 30.16\\
 \bf \textsc{mape} (\%) & 50.22 & 17.54 & 16.04 & 24.00 & 13.01

\\\bottomrule
\end{tabular}
\end{table*}

Table \ref{tab:correlation_error} shows the correlation, error values, \textsc{rae} and \textsc{mape} for each of the aforementioned algorithms that were trained with the data sets from the literature and tested with the original experimental data set. Notably, the ZeroR model predicts the mean of the analyzed values, resulting in a correlation of 0 and an error of 100 \%, which confirms that the data set is complex and that there is no predominant target class. The results improve when the \textsc{lr} model is used, resulting in a correlation of 0.73, which supports that the model is capable of detecting patterns. However, the \textsc{rae} is still very high (108.55 \%). There is no improvement in the metrics when using the \textsc{smo}reg algorithm. In the latter case, the correlation remains at 0.70, and the \textsc{rae} reaches 164.41 \%. Therefore, it can be concluded that this type of algorithm is not suitable for predicting surface roughness in material extrusion. The alternative is to opt for a tree-based model. Thus, the most basic classifier, \textsc{ds}, is selected. With this algorithm, the value of the correlation is maintained at 0.70, equal to \textsc{smo}reg, but the \textsc{rae} drops to the value obtained with the \textsc{lr} model, with an error of 107.18 \%. The best performance is obtained with the \textsc{rf} model, with the correlation value improving to 0.79, and the \textsc{mape} metric reducing to 8 \% (\textit{i.e.}, the accuracy of the model reaches 92 \%). Consequently, \textsc{rf} seems to be the best algorithm for predicting surface roughness even though the \textsc{rae} is relatively high due to the nature of this metric as indicated in Section \ref{sec:evaluation_metrics}.

\begin{table*}[!htbp]
\centering
\caption{\label{tab:correlation_error}Correlation and error values for the selected \textsc{ml} models using the data sets from the literature for training and our own experimental data set for testing (table by authors).}
\begin{tabular}{lllllll}
\toprule
 & \textbf{\textsc{z}ero\textsc{r}} & \textbf{\textsc{lr}} & \textbf{\textsc{smo}reg} & \textbf{\textsc{ds}} & \textbf{\textsc{rf}} \\\hline
 \bf Correlation & 0.00 & 0.73 & 0.70 & 0.70 & 0.79\\
 \bf \textsc{rae} (\%) & 100.00 & 108.55 & 164.41 & 107.18 & 49.88\\
 \bf \textsc{mape} (\%) & 17.08 & 18.83 & 28.70 & 17.51 & 8.00

\\\bottomrule
\end{tabular}
\end{table*}

\begin{figure*}
 \centering
 \subfloat[\centering Prediction error in training with the data from the literature]{{\includegraphics[width=7cm]{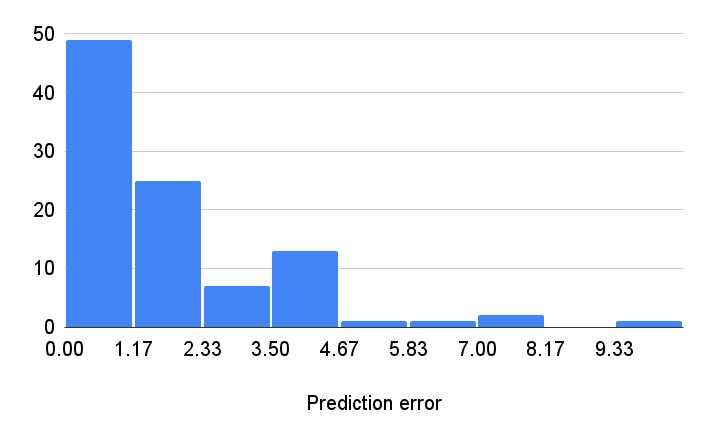}}}
 \qquad
 \subfloat[\centering Prediction error in testing with the experimental data]{{\includegraphics[width=7cm]{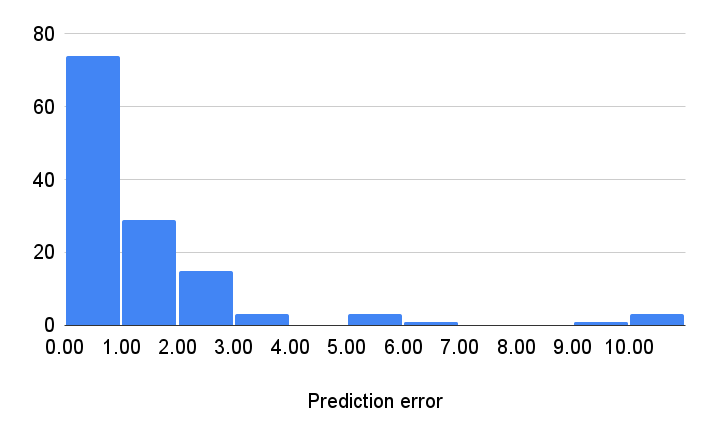}}}
 \caption{\label{fig:error_training_testing} Histogram of the prediction error for both training and testing (figure by authors).}
\end{figure*}

In the end, Figure \ref{fig:error_training_testing} shows the histogram of the prediction error for both training and testing, which corresponds to the difference between the forecast and actual value for each sample. Thus, the histogram collects the absolute errors as ranges. Note that most of the values are concentrated in lower deviation zones.

\subsubsection{Feature analysis and selection}
\label{sec:feature_analysis_selection_results}

For optimization purposes, assessing the influence of the printing parameters on the response variable is particularly interesting. Accordingly, an analysis of the significance of each selected printing parameter and three parameters available in the data set obtained from the literature (layer height, nozzle diameter, wall thickness, infill density, printing temperature, printing speed and shape) is performed using \texttt{ClassifierAttributeEval}\footnote{Available at \url{https://weka.sourceforge.io/doc.dev/weka/attributeSelection/ClassifierAttributeEval.html}, April 2023.} on the data sets gathered from the literature to evaluate the worth of an attribute by using the \textsc{rf} model. Table \ref{tab:significance} ranks these parameters based on their relevance.

\begin{table*}[!htbp]
\centering
\caption{\label{tab:significance}Significance of printing parameters (table by authors).}
\begin{tabular}{lc}
\toprule
 \textbf{Printing parameter} & \textbf{Significance}\\\hline
 Layer height & 3.15\\
 Nozzle diameter & 0.48\\
 Wall thickness & 0.24\\
 Infill density & 0.15\\
 Printing temperature & 0.01\\
 Printing speed & -0.04\\
 Shape & -0.20
\\\bottomrule
\end{tabular}
\end{table*}

Based on the results, the most significant parameter is layer height. This agrees with the general knowledge of the process \citep{Perez2018}. The nozzle diameter is the second most important factor, whereas wall thickness lies in third place. In the end, printing speed and temperature are barely relevant. However, in the experimental data, the nozzle diameter remains unchanged with a value of 0.4 mm, as previously mentioned. It should be noted that the average nozzle diameter of the training data set was 0.39 mm.

\begin{figure}
 \centering
 \subfloat[\centering Surface roughness versus layer height]{{\includegraphics[width=6cm]{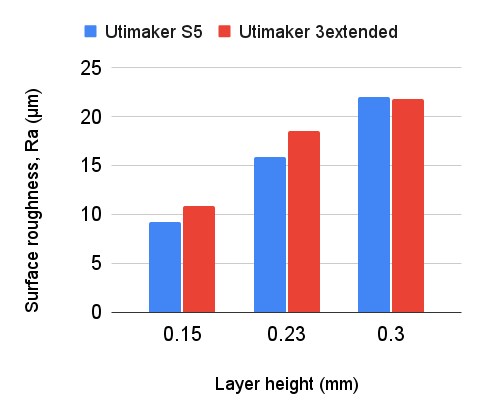}}}
 \qquad
 \subfloat[\centering Surface roughness versus wall thickness]{{\includegraphics[width=6cm]{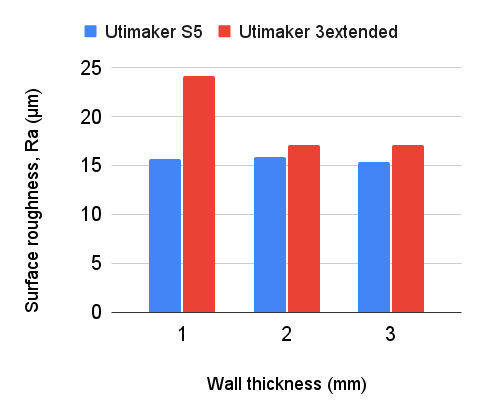}}}
 \caption{\label{fig:layer_height_wall_thickness}Surface roughness versus the layer height and wall thickness for both the Ultimaker 5S and Ultimaker 3 extended (figure by authors).}
\end{figure}

Figure \ref{fig:layer_height_wall_thickness} (a) plots the surface roughness against the layer height for the two selected 3\textsc{d} printers, demonstrating that the surface roughness increases with the layer height. The relationship between wall thickness and surface roughness is not straightforward as the latter case (see Figure \ref{fig:layer_height_wall_thickness} (b)). In both cases, the Ultimaker S5 results in lower surface roughness. Therefore, the influence of the printer is critical. Taking into account that the results of the experimental testing data depend on the printer, the same is true when dealing with the training data. This confirms that the material extrusion process lacks repeatability when it comes to surface quality.

\begin{figure}
 \centering
 \subfloat[\centering Surface roughness versus layer height]{{\includegraphics[scale=0.25]{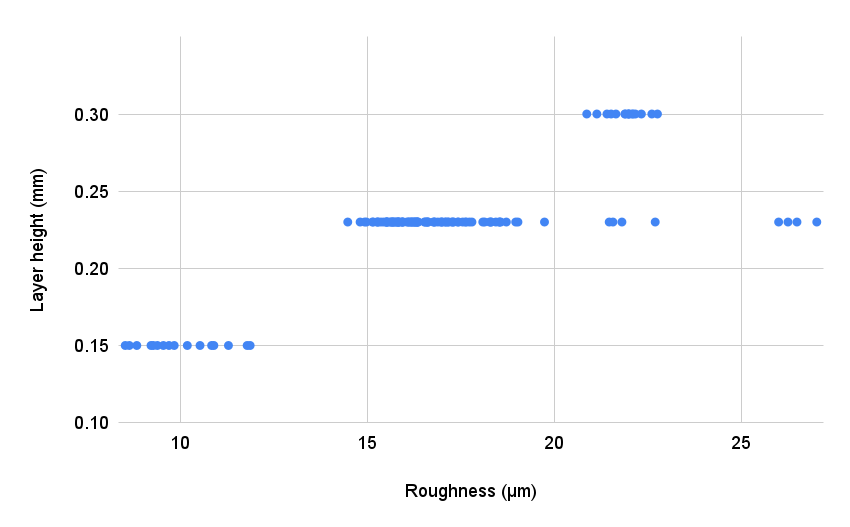}}}
 \qquad
 \subfloat[\centering Surface roughness versus wall thickness]{{\includegraphics[scale=0.25]{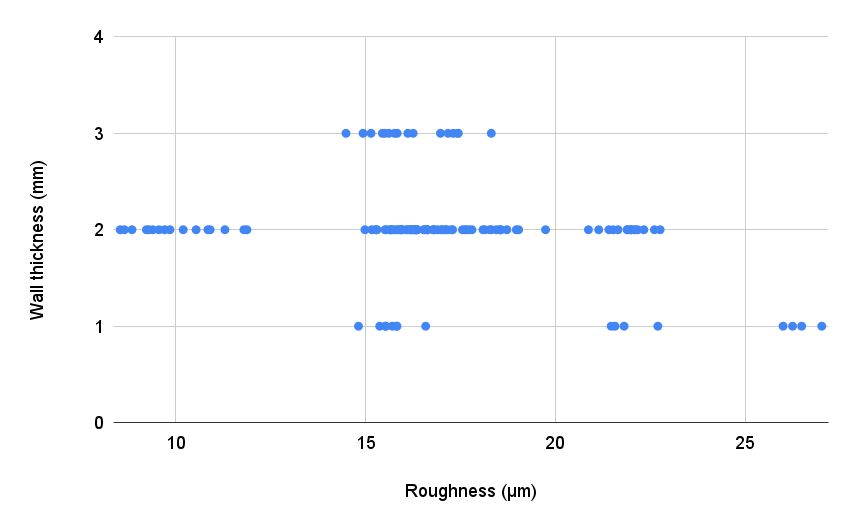}}}\\
 \subfloat[\centering Surface roughness versus printing temperature]{{\includegraphics[scale=0.25]{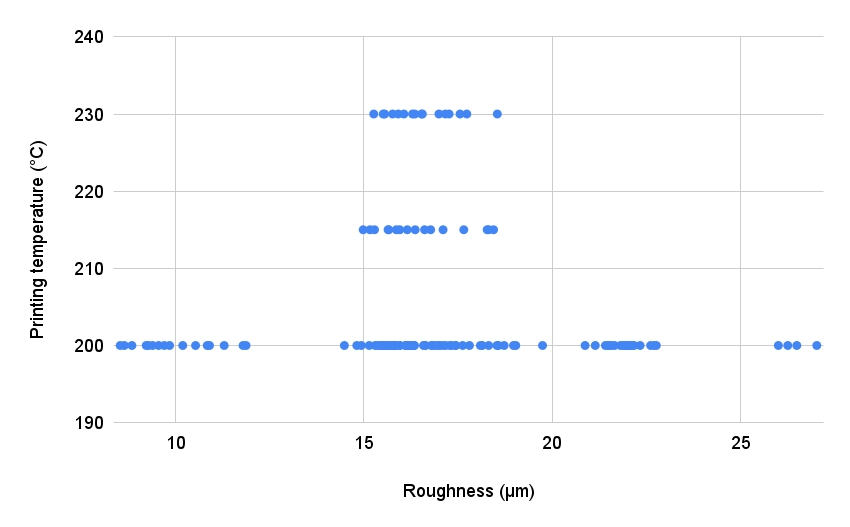}}}
 \qquad
 \subfloat[\centering Surface roughness versus printing speed]{{\includegraphics[scale=0.25]{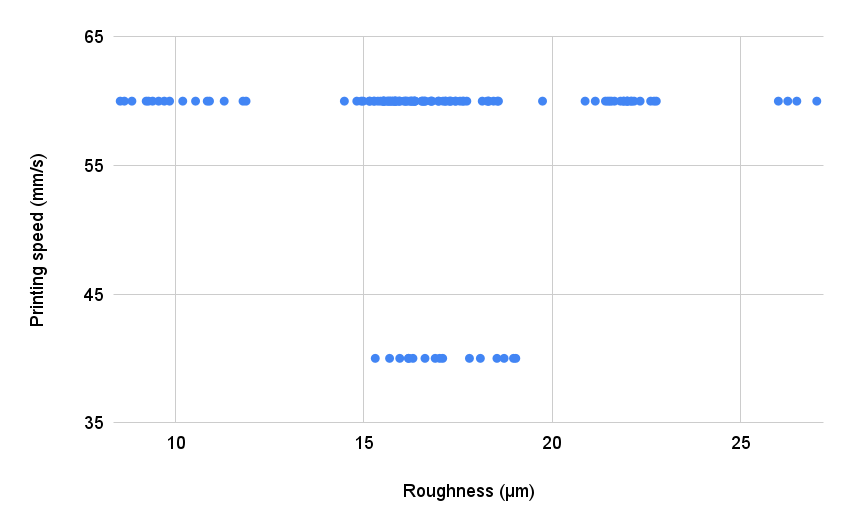}}}
 \caption{\label{fig:graphs}Surface roughness versus printing parameters for testing (figure by authors).}
\end{figure}

Figure \ref{fig:graphs} details the variation of the most significant printing parameters from the experimental data for testing from Table \ref{tab:significance}, along with the surface roughness in decreasing order of significance for the \textsc{ml} model. In light of the illustrations, the layer height is the only one that exhibits a clear, direct linear relationship with regard to surface roughness, demonstrating that it is the parameter with the highest significance. However, based on the values shown in the rest of the graphs for wall thickness, printing temperature and printing speed, there are additional and more complex relationships with surface roughness in the experimental data. Thus, a simple correlation analysis may not be effective enough for this research problem. Take the case of a surface roughness higher than \num{20} µm as a representative example. In this case, surface roughness cannot be exclusively inferred using the layer height parameter, but might feasibly be detected by an \textsc{ml} model when combined with data of wall thickness, printing temperature and printing speed.

\section{Conclusion}
\label{sec:conclusion}

The present study proposes using \textsc{ml} models to predict the surface roughness of parts printed using the material extrusion approach. As input data, this study takes advantage of previously published data and is presented as a preliminary application of this methodology. 

The results of the study help identify the following conclusions: (\textit{i}) the use of published data largely allows researchers to diminish the need to perform costly and time-consuming experiments; (\textit{ii}) \textsc{ml} algorithms have proven to be a good tool for analyzing data and building models to study surface roughness in material extrusion; (\textit{iii}) the correlations among the models were determined using only published data, and published and experimental data being 0.93 and 0.79, respectively, for the \textsc{rf} model; (\textit{iv}) when addressing the error, it should be noted that the values are relatively low (\textsc{mape} metric: training 13 \%, training and testing 8 \%); (\textit{v}) the analysis of the printing parameters' significance agrees with the existing knowledge of the process, with layer height being the most critical parameter. 

\section{Limitations and future work}
\label{sec:limitations and future work}

Certain limitations in the presented methodology for modeling surface roughness appear in a number of ways: (\textit{i}) access to data from some published papers is not available; (\textit{ii}) some measuring methodologies are not fully described, including details on the used standard, filters, cut-off and other relevant characteristics, which make it impossible to identify the measured features; (\textit{iii}) the configuration complexity of the printing process may result in significant differences among data, such as variations in the type of printer, the fan setups and the build plate, resulting in undesired variability.

In summary, despite the limitations, the proposed solution can help obtain insights from the material extrusion printing process and aid designers and researchers in optimizing this process to improve the surface quality of their products. However, future work is still needed. For instance, (\textit{i}) there is a need to elaborate larger data sets for training and testing; (\textit{ii}) additional printing factors should be included in the analysis, such as printing orientation and the influence of the machine settings on surface roughness; (\textit{iii}) the Deep Learning models can be tested; (\textit{iv}) generalization must be further researched. In this sense, both the training and the testing data sets should include different samples and contexts.

In addition, the proposed methodology should be identified as a general approach that can be further used in other manufacturing processes to predict other types of variables. The present study proves that it is possible to take advantage of the published data to improve the knowledge of the printing process with limited effort in terms of time and money investment for carrying out experiments. Although data-driven
research requires large investments in data curation, with easy access guaranteed to data for researchers and general users \citep{hey2009fourth}, there are still large opportunities to work under data-driven research in manufacturing in the coming years. 

\section*{Abbreviations}

Decision Stump (\textsc{ds}), Decision Tree (\textsc{dt}), Deep Learning (\textsc{dl}), Logistic Regression (\textsc{lr}), Machine Learning (\textsc{ml}), mean absolute percentage error (\textsc{mape}), Neural Networks (\textsc{nn}), polylactic acid (\textsc{pla}), Random Forest (\textsc{rf}), Random Tree (\textsc{rt}), relative absolute error (\textsc{rae}), Support Vector Machine (\textsc{svm}), tolerance error (\textsc{te}).

\section*{Data availability}
The data are publicly available\footnoteref{data_link}.

\section*{Acknowledgements}
This study was partially supported by the Xunta de Galicia grants ED481B-2021-118 and ED481B-2022-093, Spain.
The authors would like to thank Mr Javier Rodríguez and the Laboratorio Oficial de Metrología de Galicia for their support.

\bibliography{mybibfile}

\end{document}